\documentclass[graybox]{svmult}

\usepackage{makeidx}         % allows index generation
\usepackage{graphicx}        % standard LaTeX graphics tool
\makeindex             % used for the subject index
\newcommand{\semantics}[1]{[\![ #1 ]\!]} 
\newcommand{\ov}{\overrightarrow} 

\begin{document}

\title*{Pregroup Grammars, their Syntax and Semantics}
% Use \titlerunning{Short Title} for an abbreviated version of
% your contribution title if the original one is too long
\author{Mehrnoosh Sadrzadeh}
% Use \authorrunning{Short Title} for an abbreviated version of
% your contribution title if the original one is too long
\institute{Mehrnoosh Sadrzadeh\at Department of Computer Science, University College London \email{m.sadrzadeh@ucl.ac.uk}
}
%
% Use the package "url.sty" to avoid
% problems with special characters
% used in your e-mail or web address
%
\maketitle

\abstract{Pregroup grammars were developed in 1999 and stayed Lambek's preferred algebraic model of  grammar.  The set-theoretic semantics of pregroups, however, faces an  ambiguity problem. In his latest book, Lambek suggests that this problem might be overcome   using finite dimensional vector spaces rather than sets. What is the right notion of composition in this setting, direct sum or tensor product of spaces?}

\section{Introduction}
\label{sec:intro}

In his last published book \emph{From Rules of Grammar to Laws of Nature} \cite{Lambek2014}, on page 31, Jim Lambek says the following:

\begin{quote}
An algebraic system with linguistic applications goes back to K. Ajdukiewicz and Y. Bar-Hillel, but it is now best described as a residuated monoid.
\end{quote}

This sentence is from  the opening paragraph of  chapter 14 of the book and chapter 14 is dedicated to  Lambek's original algebraic grammar of language, other wise known as the \emph{Syntactic Calculus} \cite{Lambek58}.  On page 34, in chapter 15, Lambek revisits this historical remark with the following one:

\begin{quote}
To accommodate Miller's restriction on short term memory, we look at another algebraic system. We define a \emph{pregroup}  as a  partially ordered monoid in which each element $a$ has a \emph{left adjoint} $a^l$ and a \emph{right adjoint} $a^r$ [$\cdots$]. 
\end{quote}
 
 The rest of the chapter is dedicated to Lambek's novel algebraic grammar of language, the above mentioned \emph{pregroup} grammars. 

The trouble with pregroup grammars has always been their semantics, or lack thereof. A cut-free sequent calculus has been developed for pregroups by Buszkowski, who has also shown that the expressive power of pregroup grammars, similar to that of the Syntactic Calculus, is context-free \cite{Buszkowski-prg}. The set theoretic  semantics that one obtains for pregroups, however,  is ambiguous:  a pregroup term $abc^l$ has two semantics: $A \times C^B$ and $C^{A \times B}$. In  chapter 6, page 75,    of \cite{Lambek2008}, Lambek discusses this problem and explains that adding brackets to pregroup terms is one way of getting around it.  Later, and on page 36 of \cite{Lambek2014}, however, after  reviewing the ambiguity problem he says:

\begin{quote}
A more original interpretation of pregroup grammars has been proposed by Mehrnoosh Sadrzadeh and her collaborators. Their idea is to target the interpretations in a finite dimensional vector space  and interpreting $abc^l$ as $A + B + C^*$ where
\[
C^* = Hom(C, F)
\]
 $F$ being the underlying field. 
\end{quote}

The main purpose of this article is to go through what Lambek might have  meant in the above notation. We will do so by reviewing the syntax of pregroup grammars and then developing a vector space semantics for them. Syntactic examples are taken from chapter 3 of Lambek's first book on pregroup grammars  \emph{From Word to Sentence} \cite{Lambek2008}. A review  of the problem of ambiguity of  the semantics of pregroup grammars is then presented.  This is preceded by presenting two types of vector space semantics for pregroups. Lambek's notation when referring to these suggests that direct sum of vector spaces may be a viable solution for interpreting the monoid multiplication of pregroups. We  will show that this will not  work. We will then present a case for tensor product in place of direct sum and develop a vector semantics based on that.

Vector semantics at the word level has been developed as a successful subfield of Natural Language Processing,  see  for instance a range of developments starting from 1975 \cite{Salton,Schutze,Landauer,Turney,Lin,BullinariaLevy}. Interestingly enough,  and 
maybe not so incidentally, the origins of the ideas of the vector space models,   go back to the work of Z. Harris, who has also been referenced by Lambek in \cite{Lambek2008} and \cite{Lambek2014} as the linguist  who made use of a grammatical  notation very close to pregroup grammars. Tensorial vector space semantics for natural language has been developed in the context of what is sometimes referred to as \emph{DisCoCat}. This acronym stands for \emph{Distributional Compositional Categorical}. It uses compact closed categories and vector space instantiations thereof to develop a distributional (vectorial) semantics for phrases and sentences of natural language.  The motivating paper of this line of research already argues for the use of tensors \cite{ClarkPulman}. A categorical disposition of the model using pregroup grammars was  presented in \cite{Clark-Coecke-Sadr} and later published in Lambek's 90'th Festschrift \cite{Coeckeetal}. Non-categorical models which take advantage of multilinear algebraic notation and the use of  tensor contraction have  also been considered, see \cite{Maillard2014,Sadr2019}.  This article is in the latter category, in that we do not use  category theory and present the semantics using language from multilinear algebra. It is partly based on a talk I gave in the Federation of Logic Language Information (FoLLI) affiliated meeting  ``The Legacy of Joachim Lambek", in the 15th Congress on Logic, Methodology, and Philosophy of Science (CLMPS) in Helsinki in 2015. This article  is the first occasion  that discusses the problem of semantic ambiguity of pregroup grammars in the context of vector spaces and the first time that the  formal use of direct sum (as opposed to tensors) is considered and  investigated.

Researchers of natural language processing implement their vector space models on large corpora of data, such as the British National Corpus (BNC), and more recently, the much larger UKWaCKY, and experiment with them. The experiments  often  involve deciding about the degrees of semantic similarity or relatedness  of pairs of words, using the distances between their vector representations. In our work, we have expanded these experiments from words to sentences, and have used the distances between the vector representations of sentences to decide  the degrees of their semantic similarity. These experiments did outperform the vector models that do not consider a grammar. We showed this in a series of papers published in mainstream Natural Language Processing conferences and journals, for instance see \cite{GrefenSadr,GrefenSadrCL,kartsaklis2012,Milajevs,KartSadr}.  We will use the ideas behind these experiments and exemplify our model on a toy scenario,  built  from BNC on one of Lambek's examples of common origin of languages from \cite{Lambek2008}. We build vector and tensor semantic representations  for adjective noun phrases, sentences, and questions and use them to disambiguate the meaning of a  word of interest to Lambek, \emph{lumberjack}. This article  is  also the first instance  that semantics of questions have been considered in a \emph{DisCoCat}.

The vector spaces used in Distributional  and Compositional Distributional semantics (and their categorical models) are finite dimensional and  assumed to have a fixed orthonormal basis. We take advantage of the finite dimensionality and use the notions of direct sum and Cartesian product of finite dimensional vector spaces interchangeably.  We also take advantage of finite dimensionality and use the cosine formula to measure  the angle between the vectors and use this, as it is common practice in Distributional semantics, to formalise a notion of semantic similarity/relatedness. As the reader might note, a non-standard basis may give a very  different  set of results, but we won't go into details of this here.

%Much experimental work has been done in this framework, when one build the vectors and tensors using large coropora of data and studies the comparative performance of vectors and tensors in language tasks such as sentence similarity and verb disambiguation, 

Finally, there has been much dispute about the ability and coverage of pregroup grammars, mainly because of their semantic ambiguity problems. Lambek's opinion, however,  remains unambiguous. This is what he said on page 73, in the Epilogue chapter of \cite{Lambek2014}:

\begin{quote}
Among three mathematical formulations of syntax, I now prefer \emph{pregroups}. [$\cdots$].  If one wishes to retain a Montague type of semantics, one must abandon the uniqueness of interpretation or replace sets by finite vector spaces.
\end{quote}

\section{Pregroup Grammars}
\label{sec:2}

%\begin{svgraybox}
In the  Lambek tradition of  categorial grammars, an ${\cal X}$ grammar of a language with the vocabulary $\Sigma$  is an  algebra ${\cal X}$  freely generated by the partially ordered set of basic types ${\cal B}$ of the language and a type assignment ${\cal D}$ that assigns to element of $\Sigma$ terms from ${\cal X}$.
 In Lambek's first categorial grammar, introduced  in \cite{Lambek58},  ${\cal X}$ was a residuated monoid. Later in \cite{Lambek99},  he argued for a simplification of the setting, where ${\cal X}$ was replaced  by a pregroup. 
%\end{svgraybox}

\subsection{Mathematical Definition}
\label{sub:mathspre}

A \emph{pregroup} algebra $P$  is a partially ordered monoid $(P, \cdot, 1, \leq)$, where each element $ p \in P$ has a left and a right adjoint, denoted respectively by $p^l$ and $p^r$. These mean that the following hold:

\begin{enumerate}
\item $(P, \cdot,  1)$ is a monoid (i.e. a unital semigroup). Recall that monoids are  sets $P$ with a multiplication operator, here represented by $\cdot$, which has a unit, here represented by 1. This means that for $p,q$ elements of $P$, their multiplication $p \cdot q$ is also an element of $P$, furthermore, this multiplication has a unit element, i.e. $p \cdot 1 = 1 \cdot p = p$. Note that, however, the multiplication might not in general be commutative, i.e. in general a monoid does not have the property that $(*)\,  p \cdot q = q \cdot p$, which is desired for linguistic applications that will be discussed in this section. 
%, that is for every $p,q \in P$  we have:
%\[
%p \cdot q \in P \quad  \mbox{and}  \quad 1 \cdot p = p \cdot 1 = p.
%\]
\item $(P, \leq)$ is a partially ordered set and  the monoid multiplication of $P$  respects this partial ordering, that is for every $p,q,r \in P$ we have:
\[
p \leq q \Rightarrow p \cdot r \leq q \cdot r \quad  \mbox{and} \quad r \cdot p \leq r \cdot q
\]
\item for every $p, p^r, p^l \in P$ we have:
\[
p \cdot p^r \leq 1 \leq p^r \cdot p \quad \mbox{and} \quad p^l \cdot p \leq 1 \leq p \cdot p^l
\]
\end{enumerate} 
Given a partially ordered set of basic types of a language, denoted by  $({\cal B}, \leq)$, a type assignment  can be defined for that language as  the following relation:
\[
{\cal D} \subseteq {\cal F}({\cal B}, \leq) \times \Sigma
\] 
over ${\cal F}({\cal B}, \leq) $: the free pregroup generated by the partially ordered set  $({\cal B}, \leq)$ and $\Sigma$: the vocabulary of the language. The type assignment is also    referred to as a ``type \emph{dictionary}" or a ``\emph{lexicon}". 

Some of the properties of pregroup algebras, those often used by pregroup grammars,  are as follows.  Firstly, we often use a version of the order preservation of the monoid multiplication,  as follows:
\[
p \leq q \  \mbox{and}  \   p' \leq q'  \quad \Rightarrow   \quad p \cdot p' \leq q \cdot q' \  \mbox{and}  \ p' \cdot p \leq q' \cdot q
\]
It is sometimes necessary to take the adjoints of compound (as opposed to basic) types, to be able to compute the following inequalities: 
\[
(p \cdot q)^l  \cdot  (p \cdot q) \leq 1 \leq (p \cdot q) \cdot (p \cdot q)^l \quad \mbox{and} \quad 
(p \cdot q) \cdot (p \cdot q)^r \leq 1 \leq (p \cdot q)^r \cdot (p \cdot q)
\]
These adjoints are computed as shown below:
\[
(p \cdot q)^l = q^l \cdot p^l \quad \mbox{and} \quad (p \cdot q)^r = q^r \cdot p^r
\]
The order of the above multiplications change, since adjunction is order reversing, that is: 
\[
p \leq q \Rightarrow q^l \leq p^l \quad \mbox{and} \quad q^r \leq p^r
\]
Taking many copies of the adjoints  keeps reversing the order of the above inequalities, providing us with many copies of them and complex types, which  Lambek calls \emph{iterated adjoint types} and in effect they are the \emph{same adjoint iterated types}. The same adjoint iterated type  inequalities  have the following form:
\begin{eqnarray*}
p \leq q & \Rightarrow & \\
q^l &\leq& p^l \quad \mbox{and} \quad q^r \leq p^r\\
p^{ll} &\leq& q^{ll}  \quad \mbox{and} \quad p^{rr} \leq q^{rr}\\
q^{lll} &\leq& p^{lll} \quad \mbox{and} \quad q^{rrr} \leq p^{rrr}\\
\cdots\\
\end{eqnarray*}
Applying  the adjunction inequalities to the  same adjoint iterated types provide us with the following adjunction inequalities: 
\begin{eqnarray*}
p^{ll} \cdot p^l &\leq& 1 \leq p^l \cdot p^{ll} \qquad p^{r} \cdot p^{rr} \leq 1 \leq p^{rr} \cdot p^{r}\\
p^{lll} \cdot p^{ll} &\leq& 1 \leq p^{ll} \cdot p^{lll} \qquad p^{rr} \cdot p^{rrr} \leq 1 \leq p^{rrr} \cdot p^{rr}\\
\cdots
\end{eqnarray*} 
Mixed type  iterated adjoints also exist and  contrary to the iterated adjoint types, these cancel out, i.e. we have that:
\[
p^{lr} = p ^{rl} = p
\]
The same type iterated adjoints and their above properties are often used when there is some kind of movement in sentences and clauses, for instance they appear when analysing wh-questions.

Lambek's favourite mathematical example of a pregroup algebra is the set of order preserving  unbounded mappings on the integers $\mathbf{Z}$. This example is presented in chapter 7.2 of \cite{Lambek2008}. It consists of the set of maps $f \colon \mathbf{Z} \to \mathbf{Z}$, where for  $m,n \in \mathbf{Z}$,  the following hold, the first one defines order preservation and the second one  unboundedness:
\begin{eqnarray*}
m\leq n  &\Rightarrow & f(m) \leq f(n)\\
n \to \infty &\Rightarrow &  f(n) \to \infty
\end{eqnarray*}
The order on two maps $f,g \colon \mathbf{Z} \to \mathbf{Z}$ is defined as follows:
\[
f \leq g \Leftrightarrow f(n) \leq g(n) \quad \forall n \in \mathbf{Z}
\]
Monoid multiplication on the set of mappings  is composition of functions and  the identity function constitutes  its unit. Left and right adjoints of a function $f$ are defined as follows:
\begin{eqnarray*}
f^l(m) &=& \min \{ n \mid m \leq f(n)\}\\
f^r(m) &=& \max \{n \mid f(m) \leq n \}
\end{eqnarray*}
Since this is meant to be a linguistics contribution and there are already other papers in this volume about Lambek's mathematical contributions, we will not go through more details than this. It is, however,  a fun exercise to verify that the above definitions indeed satisfy the adjunction inequalities. 

\subsection{Linguistic Applications}
\label{subsec:linapp}

For the fragment of English considered in chapter 1 of \cite{Lambek2008}, the set of basic types and their partial orderings are discussed below. I have divided these into items and subitems according to the role they play in English grammar and the relationships between them. 
\begin{itemize}
\item  Basic Types
\begin{itemize}
\item $\pi$: subject, $o$: direct object, $s$: declarative sentence, $q$: question
\item $n$: name, $n_0$: mass noun, $n_1$: count noun,  $n_2$: plural, $\overline{n}_i$: complete noun phrase
\item $a$: predicative adjective, $\overline{a}$: predicative adjectival phrase
\item  $i$: infinitive of intransitive verb, $j$: infinitive of complete verb phrase
\item  $\pi_i$: $i$'th  person subject, in order, e.g. $\pi_1$ is the first person singular, $\pi_2$ is the second person singular, $\pi_3$ is the third person singular,  $\pi_4$ is the first person plural, and so on. 
\item $s_1, s_2, q_1, q_2$: declarative  sentence and 	question  in present or past tense, respectively, i.e. subscript 1 denotes present tense and subscript 2 past tense. 
\item $\overline{q}$: wh-question.
\end{itemize}
\item Basic Partial Orderings
\begin{itemize}
\item on subscripted types and their main type: $\pi_i \leq \pi, s_i \leq s, q_i \leq q$,
\item on nouns: $ n\leq \pi, n\leq o$, $\overline{n}\leq \pi, \overline{n}\leq o$,
\item on infinitives: $i \leq j$,
\item on questions: $q \leq \overline{q}$.
\end{itemize}
The basic partial orderings  encode the relationship between words of the types:  in general a partial ordering $x \leq y$ between two basic types $x$ and $y$ means that words that are $x$ can also be $y$, i.e., words that have grammatical type $x$ can also have grammatical type $y$. These do not include ambiguous cases, e.g. words that can be both a noun and a verb as ``chair" is, so we do not have a partial ordering such as $(*)\, n \leq v$.  Rather, they try to represent the hierarchy  between grammatical types, a very good example, which does not appear above since we do not deal with pronouns in this paper,  is the partial ordering that says a pronoun can be a subject. Examples of such hierarchical relations in the above are,   $\pi_i \leq \pi$, which says that an $i$'th person subject is a subject, $o_i \leq o$, which says that an $i$'th person object is an object, $\overline{q} \leq q$, which says that a wh-question is a question, and so on. 
\end{itemize}

Lambek did not  assign different  types to wh-questions in different tenses, i.e., he did not have a separate type for $\overline{q}_1$ and $\overline{q}_2$. In  chapter 1 and the Conclusion chapter 7.4 of \cite{Lambek2008}, however, he does explain  that these types are obtained  by ``trial and error" rather than applying a learning algorithm and that, similar to scientific postulates, they are  amenable to change in view of new evidence. So one can imagine, that if and when needed, we can enlarge our set of basic types with other desired types, including $\overline{q}_1$ and $\overline{q}_2$.  Pregroup learning algorithms have already been developed, e.g.  see the work of  Foret and Bechet  \cite{ForetBechet}.

With the above basic types and partial orderings at hand,  we  generate the pregroup algebra of our pregroup grammar and  \emph{analyse}  sentences and phrases of the above types, using our type assignment, elements of the pregroup algebra,  and axioms that govern the latter. By \emph{analysis} I mean that procedure that demonstrates how the multiplication of the types assigned to the words of a string  is in a partial ordering relation  with the type of the string. Formally speaking, given a string $\alpha$  of  words $w_1  w_2 \cdots w_n$, a set of basic types and their partial orderings,   and a  type assignment ${\cal D}$,   whenever   $\alpha$ has the grammatical type $a$,  we can show that the following partial ordering holds in the pregroup grammar:
\[
t_1 \cdot t_2 \cdot \cdots \cdot t_n \leq a \quad \mbox{for} \quad 
(w_i, t_i) \in {\cal D}\,.
\]

We use this methodology to analyse different strings of words below;  in each case the type assignment of each word is written underneath it in the following form:
\[
\begin{array}{cccccc}
w_1 & w_2 & \cdots & w_n \,.&&\\
t_1 \cdot& t_2 \cdot & \cdots  & \cdot t_n & \leq & a
\end{array}
\]
The types  and type assignments used below are from \cite{Lambek2008}. 
\begin{enumerate}
\item Sentence in present tense
\begin{itemize}
\item  She sleeps.
\[
\begin{array}{cccc}
She & sleeps.\\
\pi_3 & \pi_3^r s_1 & \leq & s_1
\end{array}
\]
\item She may sleep.
\[
\begin{array}{ccccc}
She & may & sleep.\\
\pi_3 & \pi_3^r s_1 j^l & i & \leq & s_1
\end{array}
\]
\item She sees him.
\[
\begin{array}{ccccc}
She & sees & him.\\
\pi_3 & \pi_3^r s_1 o^l & o & \leq & s_1
\end{array}
\]
\item She may see him.
\[
\begin{array}{cccccc}
She & may & see & him.\\
\pi_3 & \pi_3^r s_1 j^l & io^l  &  o & \leq & s_1
\end{array}
\]
\item She may see him tomorrow.
\[
\begin{array}{ccccccc}
She & may & see & him & tomorrow.\\
\pi_3 & \pi_3^r s_1 j^l & io^l  &  o & i^ri  & \leq & s_1
\end{array}
\]
\item She may see him in the university.
\[
\begin{array}{ccccccccc}
She & may & see & him & in & the & university.\\
\pi_3 & \pi_3^r s_1 j^l & io^l  &  o & i^r i o^l & \overline{n}_1 n_1^l &  n_1 & \leq & s_1
\end{array}
\]
\item Mary may see John.\\
\[
\begin{array}{cccccc}
Mary & may & see & John.\\
n & \pi_3^r s_1 j^l & io^l  &  n & \leq & s_1
\end{array}
\]

\item Some people may eat pork.\\
\[
\begin{array}{ccccccc}
Some & people & may & eat & pork.\\
\overline{n}_2 n_2^l & n_2 & \pi_6 ^r s_1 j^l &  io^l &  n_0 & \leq & s_1
\end{array}
\]

\item Some people are vegetarian.\\
\[
\begin{array}{cccccc}
Some & people & are & vegetarian.\\
\overline{n}_2 n_2^l & n_2 & \pi_6^r s_1 \overline{a}^l &  a & \leq & s_1
\end{array}
\]

\item The tall old woman ate vegetables.
\[
\begin{array}{cccccccc}
The & tall & old & woman & ate & vegetables.\\
\overline{n}_1 n_1^l & n_1n_!^l & n_1 n_1^l & n_1 & \pi_3^r s_2 {o}^l &  n_2 & \leq & s_2
\end{array}
\]
\item The old woman ate vegetables and rice.\\
\[
\begin{array}{ccccccccc}
The  & old & man & ate  & vegetables & and & rice.\\
\overline{n}_1 n_1^l & n_1n_1^l  & n_1 & \pi_3^r s_2 \overline{o}^l &  n_1 & n_1^r o \, n_0^l & n_0 & \leq & s_2
\end{array}
\]
\item The old woman slept and snored.
\[
\begin{array}{cccccccc}
The & old & woman & slept & and & snored.\\
\overline{n}_1 n_1^l & n_1n_1^l  & n_1 & \pi_3^r s_2 & (\pi_3^r s_2)^r (\pi_3^r s_2) (\pi_3^r s_2)^l &  \pi_3^r s_2 & \leq & s_2
\end{array}
\]

\end{itemize}
\item Yes-No Question in present tense
\begin{itemize}
\item May she sleep?
\[
\begin{array}{ccccc}
May & she & sleep?\\
q_1 i^l \pi^l & \pi_3 & i & \leq & q_1
\end{array}
\]
\item May she see him?
\[
\begin{array}{cccccc}
May & she & see & him?\\
q_1 i^l \pi^l & \pi_3 & io^l & o & \leq & q_1
\end{array}
\]
\end{itemize}
\item Wh-Question in present tense
\begin{itemize}
\item Who  may sleep?
\[
\begin{array}{ccccc}
Who & may & sleep?\\
\overline{q} s_1^l \pi_3 &\pi_3^r s_1i^l & i& \leq& \overline{q}
\end{array}
\]
\item Who may eat pork?
\[
\begin{array}{cccccc}
Who & may & eat & pork?\\
\overline{q} s_1^l \pi_3 & \pi_3^r s_1i^l   & io^l & n_0 & \leq& \overline{q}
\end{array}
\]
\item Who  ate vegetables?
\[
\begin{array}{ccccc}
Who &  ate & vegetables?\\
\overline{q} s_2^l \pi_3 & \pi_3^r s_2 o^l& n_2 & \leq& \overline{q}
\end{array}
\]

\item Whom may she see?
\[
\begin{array}{cccccc}
Whom & may & she & see?\\
\overline{q} o^{ll}q^l & q_1j^l\pi^l & \pi_3 & io^l & \leq& \overline{q}
\end{array}
\]
\item When  may she see him?
\[
\begin{array}{ccccccc}
When  & may & she & see & him ?\\
\overline{q} i^l i^{ll} q^l & q_1j^l\pi^l & \pi_3 & io^l & o &  \leq& \overline{q}
\end{array}
\]
\item Where  may she see him?
\[
\begin{array}{ccccccc}
Where  & may & she & see & him ?\\
\overline{q} i^l i^{ll} q^l & q_1j^l\pi^l & \pi_3 & io^l & o &  \leq& \overline{q}
\end{array}
\]
\item What did the old man  eat?
\[
\begin{array}{cccccccc}
What & did & the & old & man & eat?\\
\overline{q}o^{ll}q^l & q_1j^l\pi^l & \overline{n}_1 n_1^l & n_1n_1^l  & n_1 & io^l & \leq & q_2
\end{array}
\]
\end{itemize}
Iterated adjoints $o^{ll}$ and $i^{ll}$  show up in the above wh-question that are about the objects or prepositional phrases of the sentences.  These types of questions are canonical examples of the \emph{movement} phenomena in language. It was in order to analyse these phenomena that Chomsky introduced the notion of \emph{trace}. A trace is a blank marker, usually denoted by $-$, which replaces the original location of a word in a phrase/sentence. Chomsky believed that many of the linguistic phenomena resulted from a change of location, or a \emph{movement}, of words. The most well known examples of the use of traces are in modelling wh-questions. For instance, consider the first question below ``Whom may she see?".  According to Chomsky, this question resulted from moving the object in an original sentence, such as ``She may see a man", where ``man", i.e.  the object  of the sentence, has moved from the end of the sentence to the beginning of the sentence and subsequently  replaced by the wh word ``whom". The original location of what ``whom" is referring to is marked with a $-$ in the wh-question, as depicted below by  the blank marker $-$. Lambek believed that  iterated adjoints show up in the pregroup grammar exactly in those places where a trace was used by Chomsky. Rightly so, and as analysed above, the pregroup type of ``whom" is $\overline{q}o^{ll}q^l$, which has a same type iterated adjoint $q^{ll}$. We have presented the types of the wh words  above and in what follows  present the corresponding traces:
\begin{itemize}
\item Whom may  she  see  $-$   ?
\item When may she see him $-$ ?
\item Where may she see him $-$ ?
\item What did the old man eat $-$ ?
\end{itemize}
In the cases where there is no movement, e.g.  when the wh-question is  about the subject of the sentence , e.g. in: \emph{Who may sleep? Who may eat pork? Who ate vegetables?}, there are no iterated adjoints in the type of the wh-word. Rightly so, and again as analysed above, the wh word ``who" in wh-questions ``Who may sleep", ``Who may eat pork" and so on has type e.g. $\overline{q}s_1^l \pi_3$, which does not have any iterated adjoints. Iterated adjoints  show up in other movement phenomena as well, such as relative clauses. These  can be subjective or objective, wh-word used in the subjective clauses have types without iterated adjoints, wh-words in objective clauses have types with iterated adjoints. In the subjective case, no movement happens, only the wh-word ``who" substitutes where the subject was in the original sentence.  In the objective case, we witness the movement of the object from the end of the original sentence to the beginning of the relative clause. An example of each case is presented below.
\begin{itemize}
\item Men who saw John. 
\[
\begin{array}{ccccc}
Men & who & saw & John&\\
n_2  & n^r \overline{n} s^l o & n & \leq & \overline{n}\\
\end{array}
\]
\item Men whom John saw. : Men whom John saw $-$.
\[
\begin{array}{ccccc}
Men & whom & John & saw\\
n_2 & n^r \overline{n} n^{ll} s^l & n& \pi_3^r s_2 o^l & \leq \overline{n}
\end{array}
\]
\end{itemize}
\end{enumerate}

When defining a pregroup $P$, we asked for the left and right adjoints of each type $p \in P$  to satisfy two pairs of  inequalities: a \emph{contraction} pair:
\[
p^l  \cdot p \leq 1  \quad \mbox{and} \quad p \cdot p^r \leq 1
\]
and an \emph{expansio}n pair:
\[
1 \leq p \cdot p^l \quad \mbox{and} \quad 1 \leq p^r \cdot p 
\]
You might have noticed that none of the computations in our above examples need any of the expansion inequalities. Lambek justifies this in chapter 7.2 of \cite{Lambek2008} by proving a \emph{switching} lemma, as follows:
\begin{quote}
When proving that $x \leq y$ in the free pregroup ${\cal F}({\cal B}, \leq)$, one may assume, without loss of generality, that all (generalized) contractions precede all (generalized) expansions. Hence, if $y$ is a simple type, no expansions are needed.  
\end{quote}
Later, Preller showed that expansions are needed when  going beyond syntax and getting involved in analysing  semantics of sentences  \cite{Preller07,Preller11}.  In our work on vector space semantics,  we have  also often used Preller's work and expansion maps, e.g. see  the semantics of negation  \cite{PrellerSadr},  relative pronouns  \cite{RelPronMoL}, coordinators \cite{kartsaklisphd},  and quantifiers \cite{HedgesSadr2019}.

\section{Set Theoretic and Vector Space Semantics}
\label{sec:prgsem}

Chapter 6 of \cite{Lambek2008} demonstrates how a set theoretic semantics for pregroup grammars, developed  in the style of Richard Montague  for phrase structure grammars  \cite{Montague1970}, leads to  \emph{ambiguity}. We  go through this problem below and then show how by moving from sets to vector spaces one might overcome it. 

\subsection{Ambiguous Set Theoretic Semantics}
\label{subsec:setsem}

In order to assign a set theoretic semantics to a pregroup (or any) grammar of a language,  one first assigns sets to basic types of the language,  then defines set theoretic counterparts for operations of the underlying  pregroup algebra of the grammar.  For simplicity, and following Lambek, suppose we work in a pregroup grammar whose set of basic types has only two elements: names and sentences, that is, we have  ${\cal B} = \{n,s\}$. We assign the set $N$ to the type $n$ and the set $S$ to the type $s$, that is: 
\[
\semantics{n} := N, \semantics{s} := S
\]
Assuming that $N$ is the set of human beings and  $S$ is the set of truth values, this assignment expresses the fact that words with type $n$, i.e. names, are elements of $N$ and words with type $s$, i.e. sentences, are elements of $S$

For simple types $x, y$ of  our  pregroup grammar ${\cal F}({\cal B}, \leq)$,  we define the semantics of $x \cdot y$ to be the Cartesian product of the semantics of $x$ with  the semantics of $y$, that is:
\[
\semantics{x \cdot y} := \semantics{x} \times \semantics{y}
\]
When adjoint types are involved in the multiplication, function spaces are used to assign  semantics to the multiplied type, that is we define:
\[
\semantics{x^r \cdot y} := \semantics{y}^{\semantics{x}} \qquad 
\semantics{y \cdot x^l} := \semantics{y}^{\semantics{x}}
\]
where $\semantics{y}^{\semantics{x}}$ is the set of functions from $\semantics{x}$ to $\semantics{y}$.  There is   a sense of dissatisfaction around this semantics, as it is not compositional:  
\begin{eqnarray*}
\semantics{x^r \cdot y}  &\neq& \semantics{x^r} \times \semantics{y}\\
\semantics{y \cdot x^l} &\neq& \semantics{y} \times \semantics{x^l}
\end{eqnarray*}
This is so, because we have not defined a semantics for the left and right adjoints of types independently from their multiplications.  That is we have not specified  $\semantics{x^l}$ and  $\semantics{x^r}$. 

This problem manifests itself in the following example of Lambek from \cite{Lambek2008}.  Consider the type $x\cdot y\cdot z^l$ of a pregroup grammar. This type can have two different semantics,  depending on how we bracket it. If we bracket it as $(x\cdot y)\cdot z^l$, its semantics will be the set of functions from $\semantics{z}$ to $\semantics{x\cdot y}$, that is:
\[
\semantics{(x\cdot y)\cdot z^l} = \semantics{x\cdot y}^{\semantics{z}} = (\semantics{x} \times \semantics{y})^{\semantics{z}}
\]
Whereas, if we bracket it the other way around, that is as $x\cdot (y\cdot z^l)$, its semantics will become the cross product of $\semantics{x}$ with the set of functions from $\semantics{z}$ to $\semantics{y}$,  that is
\[
\semantics{x \cdot (y\cdot z^l)} = \semantics{x} \times (\semantics{y}^{\semantics{z}})
\]
The above two semantics are not equal, and as a result, the pregroup term $x\cdot y\cdot z^l$ will be assigned  two different semantics. This, expressed below, is what Lambek calls \emph{ambiguity} of set theoretic semantics and is undesirable. 
\[
x\cdot y\cdot z^l = (x\cdot y)\cdot z^l = x\cdot (y\cdot z^l),  \quad \mbox{however} \quad \semantics{(x\cdot y)\cdot z^l} \neq \semantics{x\cdot (y\cdot z^l)}
\]

One way to get around this problem, Lambek suggests, is to take the brackets of the pregroup types seriously and do not equate  $(x\cdot y)\cdot z^l$ with $x\cdot (y\cdot z^l)$. This means that we cannot have a non bracketed term such as $x\cdot y\cdot z^l$ in our pregroup grammar, unless  we agree on a default bracketing convention, e.g. that  multiplication takes precedence over adjoints.  As Lambek diagnoses it, however, pregroups are by definition associative and introducing brackets in them will need the introduction of a new mathematical object:  for example, one  whose underlying algebra is a non associative structure.  One problem  with this is that we will lose our mathematical example of  a pregroup algebra, i.e. the set of order preserving unbounded functions on the integers, since function composition is  associative.

The above problem can be approached differently. Had we defined a compositional semantics for adjoint types as explained above, we would have had
\[
\semantics{x\cdot y\cdot z^l} = \semantics{x} \times \semantics{y} \times \semantics{z^l}
\]
This would be the same,  no matter how we bracketed $x\cdot y\cdot z^l$, since we would have had: 
\begin{eqnarray*}
\semantics{(x\cdot y)\cdot z^l} &=& \semantics{x\cdot y} \times \semantics{z^l} = (\semantics{x} \times \semantics{y}) \times \semantics{z^l}  \\
\semantics{x\cdot (y\cdot z^l)} &=&  \semantics{x} \ \times \semantics{y\cdot z^l} = \semantics{x} \times (\semantics{y} \times \semantics{z^l})  
\end{eqnarray*}
The above would  have   been  equivalent, as cartesian product of sets is associative
(up to bijection):
\[
(\semantics{x} \times \semantics{y}) \times \semantics{z^l}  
\cong
\semantics{x} \times (\semantics{y} \times \semantics{z^l}) 
\cong
\semantics{x} \times \semantics{y} \times \semantics{z^l}
\]

\subsection{Vector Space Semantics}
\label{subsec:vectsem}
If instead of sets, we work with vector spaces  and send the monoid multiplication of types to the tensor product of vector spaces, the ambiguity problem will be resolved. Confining ourselves to the rudimentary pregroup grammar of the previous subsection and finite dimensional vector spaces, this  means that we work with two finite dimensional vector spaces $\mathbf{N}_\mathbf{k}$ and $\mathbf{S}_\mathbf{k}$ over a field $\mathbf{k}$. Semantics of  basic types $n$  and $s$ will now be these two spaces respectively:
\[
\semantics{n} := \mathbf{N}_{\mathbf{k}}, \quad \semantics{s} :=\mathbf{S}_{\mathbf{k}}
\]
Semantics of a multiplication of types is the tensor product of semantics of each multiplicand:
\[
\semantics{x\cdot y} := \semantics{x} \otimes \semantics{y}
\]
Adjoint types are assigned  the dual spaces of the semantics of their underlying types:
\[
\semantics{x^l} = \semantics{x^r} := \semantics{x}^*
\]
We recover a \emph{linear function space} version of the the function spaces of the set theoretic semantics. This is due to the tensor-hom relationship, i.e. there is a natural
map
\[
V^* \otimes W \to Hom(V,W)
\]
where  $Hom(V,W)$ is the set of \emph{linear maps} from $V$ to $W$ and $V^*$ is the dual space of $V$, that is the space of linear functionals from $V$ to the underlying field, defined as follows:
\[
V^* := Hom(V, \mathbf{k})
\]
For $v \in V, w \in W$ and $\alpha \in Hom(V, \mathbf{k})$, the tensor-hom relationship is concretely given as follows:
\[
(f(\alpha \otimes w))(v) := \alpha(v) \cdot w 
\]
 When $V$ and $W$ are finite dimensional, the tensor-hom relationship  becomes an isomorphism,  that is:
\[
V^* \otimes W \cong Hom(V,W)
\]
Using tensor-hom and our semantics of the left and right adjoints, we obtain the following semantics for the compound pregroup types: 
\begin{eqnarray*}
\semantics{y\cdot x^l} &:=&Hom(\semantics{x},  \semantics{y}) \cong \semantics{x}^* \otimes \semantics{y}\\
\semantics{x^r\cdot y} &:=& Hom(\semantics{x}, \semantics{y}) \cong \semantics{x}^* \otimes \semantics{y}
\end{eqnarray*}
Now consider again the problematic type $x \cdot y \cdot z^l$ and compute the semantics of its two bracketing options, as follows:
\begin{eqnarray*}
\semantics{(x \cdot y) \cdot  z^l} &=& \semantics{x \cdot y} \otimes \semantics{z^l} = Hom(\semantics{z}, \semantics{x} \otimes \semantics{y}) =   \semantics{z}^* \otimes (\semantics{x} \otimes \semantics{y} )\\
\semantics{x \cdot (y \cdot z^l)} &=&  \semantics{x} \otimes \semantics{y \cdot z^l} =  \semantics{x} \otimes Hom(\semantics{z}, \semantics{y}) =  \semantics{x} \otimes   (\semantics{z}^* \otimes \semantics{y}) 
\end{eqnarray*}
The above are equivalent, since the  tensor of finite dimensional vector spaces is symmetric and associative (up to isomorphism): 
\[
 \semantics{z}^* \otimes \semantics{x} \otimes \semantics{y}  \cong 
 \semantics{x} \otimes   \semantics{z}^* \otimes \semantics{y} 
 \]

\subsection{Direct Product of Vector Spaces}
\label{subsec:directsum}

Let us reconsider Lambek's quotation from page 36 of \cite{Lambek2014} and focus on its second half, where after mentioning vector spaces, he says:
\begin{quotation}
$\cdots$ \ and interpreting $abc^l$ as $A + B + C^*$ where
\[
C^* = Hom(C, F)
\]
for $F$ the underlying field. 
\end{quotation}

Although Lambek does not explicitly say ``finite dimensional vector spaces" here, based on an earlier quote on page 36 of \cite{Lambek2014}, which was discussed above, we are sure these are the vector spaces he meant. Another option may be  actual finite vector spaces,  the only examples of which are finite dimensional vector spaces over finite fields, e.g. finite dimensional vector spaces over $Z_p$, the integers modulo a prime $p$. At first sight, these do not seem appropriate for linguistic applications, as computations that yield word and thus phrase/sentence vector  representations take values from real numbers. One can, however, envision approximating these with an upper bound. Studying consequences and potential usefulness of these models is a future direction the author might pursue.

Here, Lambek is suggesting to use  dual spaces to interpret the adjoint types of a pregroup grammar, but his suggestion for the monoid multiplication seems to be different from ours: ``$A + B + C^*$"  is used in a place where we would normally use  ``$A \otimes B \otimes C^*$". In the category of finite dimensional vector spaces and linear maps,   $V \times W \cong V + W$, and this is what  we believe Lambek is suggesting. 
%Since we work with finite dimensional vector spaces, there is no such thing as finite direct sums of vector spaces (there is only the notion of a finite direct sum of subspaces of a  space). 
One can form the direct (Cartesian) product $A \times B \times C^*$, which is again a vector space.  Since  $A \times B \times C^*$ satisfies the universal property of a coproduct, then it could be written as $A + B + C^*$.     
%Note that in infinite dimensional, one can form the direct sum of vector spaces and this is denoted by $V \oplus W$, 

In this paper our focus is on finite dimensional vector spaces. Infinite dimensional vector spaces do not naturally arise in distributional semantics,. One can, however,  still consider them in a slightly  different setting, e.g. when we do not fix a corpus of a document and instead work with all the potential documents that can ever be written. In such a set up,  we have to work with  an infinite  set of contexts and thus our vector spaces will have to have infinitely many  basis vectors: the contexts in which a word can in principle occur. The documents, and/or contexts of this setting  are in principle infinite. We defer this to future work. 

In this section, we use  the isomorphism $V \times W \cong V + W$, work with direct product as $+$, and ask  whether one can use the direct product of vector spaces  instead of  their tensor product to obtain a non ambiguous semantics for pregroups.

Recall the definitions of the tensor and direct sum operations. A tensor product $V \otimes W$ of two spaces $V$, spanned by basis $\{v_i: i \in I\}$, and $W$, spanned by $\{w_j: j \in J\}$,  is a space spanned by the following basis:
\[
\{v_i \otimes w_j: (i,j) \in I \times J\}
\]
where $v_i \otimes w_j$ is %notation for a 
the image of $(v_i,w_j)$ under the canonical
bilinear map \\ $- \otimes - \colon V \times W \to V \otimes W$.  An element $v \otimes w$ of $V \otimes W$ is called the tensor product of $v$ and $w$. Not all elements of $V \otimes W$ are of this sort, i.e. there are elements $x$ in $V \otimes W$ that cannot be written as the tensor product of a vector from $V$ and a vector from $W$. The former elements $v \otimes w$ are sometimes referred to as \emph{pure tensors}. One of the main characteristics of finite dimensional tensor spaces
 is the tensor-hom isomorphism described and used in the previous subsection.  From now
on, we assume all spaces are finite dimensional.

A direct sum $V + W$ (in Lambek's notation) of (the same as above) $V$ and $W$  has the following basis: 
\[
\{(v_i, 0), (0,w_j): i \in I, j \in J\}
\]
The basis of the tensor and direct sum spaces  justify the slogan that \emph{tensor product of vector spaces are to Cartesian product of sets as direct sums of vector spaces are to disjoint unions of sets}.  In order to see this better, observe that the basis of a tensor product of spaces is bijective with:
\[
\{v_i: i \in I\} \times \{w_j: j \in J\}
\]
whereas the basis of a direct sum of spaces is bijective with: 
\[
\{v_i: i \in I\} \uplus \{w_j : j \in J\}
\]
where $\times$ and $\uplus$ are the Cartesian product and disjoint union of sets, respectively.\footnote{Despite this notational analogy and the slogan above, the direct sum of two finite dimensional vector spaces has the same dimension as the Cartesian product of them, and thus the two are isomorphic. 
%Note, however, that in the case of the tensor space, we are talking about Cartesian %product of basis, whereas for the direct sum, the Cartesian product is on the vectors %themselves. There is no bilinearity constraint on the Cartesian product of vector spaces, %so this is a  space of concatenated vectors of each space, whereas elements of the %tensor space are bilinear maps.  In terms of the number of vectors in a vector space,
But while the Cartesian product space might seem huge, in terms of dimensions, this is not such a  huge space, as its dimensions add.  The tensor product, on the other hand, can be seen as huge, since its dimensions multiply.}  bv

\subsection{Two Attempts}
\label{subsec:fix}
Our first attempt for using the direct sum of (finite dimensional) vector spaces is to adhere to Lambek's original method. We start by   interpreting the multiplication of simple pregroup types as the direct sum of their vector space interpretations. That is, 
\[
\semantics{x \cdot y} := \semantics{x} + \semantics{y}
\]
Moving forward,  we  keep the \emph{linear function space} interpretation for the semantics of adjoint types (rather than using the dual space interpretation of adjoints directly). That is, we  still interpret $x^r \cdot y$ and $y \cdot x^l$ as the set of linear maps from $\semantics{x}$ to $\semantics{y}$. Let us now use this interpretation and  compute the semantics of the problematic type $x \cdot y \cdot z^l$ . Semantics of the first bracketing of this types is computed as follows:
\begin{eqnarray*}
\semantics{(x \cdot y) \cdot z^l} &=& Hom(\semantics{z}, \semantics{x}+\semantics{y})\\
&=& \semantics{z}^* \otimes (\semantics{x} + \semantics{y})
\end{eqnarray*}
This is, however,  not (in general) the same as the semantics of the second way of bracketing it, computed as follows:
\begin{eqnarray*}
\semantics{x \cdot (y \cdot z^l)} &=& \semantics{x} + Hom(\semantics{z},\semantics{y})\\
&=& \semantics{x} + (\semantics{z}^* \otimes \semantics{y})
\end{eqnarray*}
To be precise (and as argued beautifully by our reviewer), if $dim(\semantics{x}) = m$,
$dim(\semantics{y}) = n$, $dim(\semantics{z}) = p$, then
$dim( \semantics{(x \cdot y) \cdot z^l}) = p(m+n)$, whereas 
$dim(\semantics{x \cdot (y \cdot z^l)} = m+(pn)$. Notice for positive integers,
$p(m+n) = m + (pn)$ implies
  $p = 1$. So the equation fails if $p > 1$.
So  we  run into the same ambiguity problem as in the set theoretic semantics.

In the above, we used part of the  direct sum semantics and part of the tensor product semantics. The latter was via the \emph{linear function} interpretation of compound adjoint types. This enabled us to apply the tensor-hom duality to the linear function spaces but stopped us from obtaining a fully compositional semantics where the adjoint types $x^r, x^l$ get a direct interpretation. 
A second possibility towards finding a fix,  is to let go of the tensor product completely and only use the direct sum, that is, we interpret the multiplication of pregroup types as the direct sum of their vector space interpretations:
\[
\semantics{x \cdot  y} := \semantics{x} + \semantics{y}
\]
The adjoint types are interpreted  as the dual spaces, as before:
\[
\semantics{x^r} = \semantics{x^l} := \semantics{x}^*
\]
 In this case, we will get a fully compositional semantics and the problem with the ambiguity  will get resolved as well. This is computed below; note that   similar to  tensor product of spaces,  direct sum  is also associative:
 \begin{eqnarray*}
 \semantics{(x \cdot y) \cdot z^l} &=& (\semantics{x} + \semantics{y}) + \semantics{z}^*\\
 \semantics{x \cdot (y \cdot z^l)} &=& \semantics{x} +  (\semantics{y} + \semantics{z}^*)
 \end{eqnarray*}
 This seems to be a reasonable solution, except that we do not have an intuitive interpretation for function types.  Semantics of functional pregroup types $x^r \cdot y$ and $y \cdot x^l$ become as follows
 \[
 \semantics{x^r \cdot y} = \semantics{y \cdot x^l} = \semantics{x}^* + \semantics{y}
 \]
We do not, however,  have a direct sum-hom duality. A corresponding equivalence with direct sum instead of tensor  fails, that is:
\[
Hom(V,W) \cong \!\!\!\!\!\! /  \ V^* + W
\]
The main reason is that  the dimensions of the left and the right hand sides are not the same. If we denote the dimensions of a vector space $X$ by $dim(X)$, then we have $dim(V) \times dim(W)$ for  the dimensions of the vector space on the left, whereas the dimensions of the right hand side vector space is  rather $dim(V) + dim(W)$. 

In \cite{Clark2013}, Clark  also considers the direct sum of vector spaces as an alternative to their tensor product. His argument in favour of a direct sum space is exactly its low additive dimensionality,  in contrast with the high multiplicative dimensions  of a tensor product space. He, however, rules the direct sum out, due to its inability to encode the \emph{interaction} between different parts of the space. He argues that an element of a direct sum space can be written down as a sum of elements of each space, with each summand encoding a separate set of properties. An element of a tensor space, however, can in general not be separated into a tensor product of two elements, giving it a chance to  model the interactions between properties of different parts of the space.  

%What if we work with direct sums of spaces that have disjoint basis? In this case, the following holds about the basis of the direct sum space:
%\[
%\{u_i: i \in I\} \uplus \{u_j: j \in J\} \cong \{u_i: i \in I\} \times \{u_j: j \in J\} 
%\]
%and since we have the tensor-hom duality for the space spanned by the right hand side set, we will also have it for the space spanned by the left hand side space.  But this means that for the rudimentary pregroup grammar considered above, we should have the  two spaces $\mathbf{N}_{\mathbf{k}}$ and $\mathbf{S}_{\mathbf{k}}$ take basis from two disjoint sets. We can do so, by limiting the first one to  the $n$-dimensional space of all words of the vocabulary and the second one to the 2-dimensional spaces of truth and falsity, or 1 and 0. 

\section{ Data-Driven Vector Space Semantics}
\label{sec:data}

After introducing  pregroup grammars, we  exemplified one on  a fragment of English. In this section, we show how to construct a vector space semantics for some of those examples.  We start by introducing the vector space semantics of words, using   text from Lambek's 2008 book \cite{Lambek2008}. The book starts with an introduction on origins of language, where Lambek provides evidence that supports the  belief that ``all languages spoken today are descended from one protolanguage". Amongst  Lambek's examples are the English \emph{father, mother, son, daughter} and the Sanskrit $pitr, m\bar{a}tr, s\bar{u}nu$, and \emph{duhitr}. To this, we add evidence from Persian: ${pedar, m\bar{a}dar, pesar}$, and \emph{dokhtar}. The only word that differs here is the one for \emph{son}, which is \emph{pesar}. The resemblance between Persian and Sanskrit is less surprising than between English and Sanskrit, since Persian is a direct descendant  of Sanskrit. 

Lambek mentions that tracing  related words of different languages is not easy, since ``words combine sound and meaning and both may change over time". As an example, he goes through the etymology of the word \emph{lumberjack}. 

\begin{quote}
The first part lumber may be recognized in long beard,
but the meaning was transformed in a relatively short time. The Germanic tribe of the long
bearded Langobardi settled in Italy as Lombards. Lombards became bankers in Renaissance
Europe; but loan institutions often degenerated into pawnshops, which ultimately also sold
second-hand furniture. Discarded furniture was stored in the lumber room, which later also
contained other pieces of wood. 
\end{quote} 

Ideas of Firth  \cite{Firth} and of Harris \cite{Harris}, that words that often occur in the same context have similar meanings,  have led to the development of a vector space model of meaning for words. Herein, one fixes a set of words of language and considers them as \emph{context words}. A context  is then defined to be a neighbourhood  window of $k$ words (e.g.  k = 5) around a word.  Given a set of target words, one builds a co-occurrence matrix from a corpus, by counting how many times a target word has occurred in the context of a context word; see \cite{Rubenstein} for one of the first places where such a matrix was constructed.  The raw count entries of the matrix are often normalised by a log likelihood function, such as Pointwise Mutual Information, for a formal definition and more normalising functions, see   \cite{Evert}. A vector representation for target words is obtained by embedding each row of this matrix into a vector space spanned by its columns. 

We build a vector representation for \emph{lumberjack}, hoping to find co-occurrences with some of the words that may relate to its origins. We use the British National Corpus (BNC) as our first source. There are 36 occurrences of the word  \emph{lumberjack} in BNC. Amongst these, there are 3 co-occurrences with \emph{wood}, 3 with \emph{log}, 5 with \emph{saw} (as in sawing wood), and 5 with \emph{tree}. There are 10 co-occurrences with \emph{shirt} and 2 with \emph{boot}. The example sentences are from different sources. Some denote the wood-related meaning of \emph{lumberjack}:

\begin{quote}
Charlesworth and Nathan (1982) tell a very poignant story about a young man who always wanted to be a lumberjack. He wandered up to the logging camp on his eighteenth birthday and enthusiastically asked for a job. (source: ``Anxiety and stress management". Enright, Simon and Powell, Trevor. London: Routledge \& Kegan Paul plc, 1990)
\end{quote}

Some are analogical and about a dog called \emph{Lumberjack}, in order to strengthen the analogy, there are many co-occurrences of this \emph{Lumberjack} with the wood-related meaning of the word:
\begin{quote}
Somewhere down the hill Lumberjack began to bark. India-May had called him Lumberjack because his bark sounded just like someone sawing wood. 

India-May locked Lumberjack inside the house. As Nathan pulled away for the second time he could hear Lumberjack in the kitchen, frantically sawing the legs off tables and chairs. 

(source: ``The five gates of hell". Thomson, Rupert. London: Bloomsbury Pub. Ltd, 1991)
\end{quote}

\noindent
Other co-occurrences  are about an emerging fashion related meaning:

\begin{quote}POP'S most flamboyant dresser -- Elton John meets -- rock's dowdiest -- Bryan Adams. But Elton probably wouldn't mind exchanging his sparkling shorts to suffer Adams' lumberjack shirts if it meant having a song at number one for 16 weeks like the Canadian rocker did BITTER BATTLE. (source: The Daily Mirror)
\end{quote}

There is no co-occurrence, however,  with \emph{beard}, \emph{pawnshop} (or \emph{pawn}), \emph{loan}, \emph{bank}, or \emph{furniture}.  When we add the Wikipedia article entry on \emph{lumberjack} to the corpus, we acquire more occurrences with wood-related words, such as \emph{log} and \emph{tree}. We also get more co-occurrences with   fashion-related words, such as \emph{shirt, boot}, in which context  the word  \emph{beard} soon shows up as well. The Wikipedia entry mentions that  ``the term lumberjack is of Canadian derivation". The wood-related  meaning of the word is the most emphasised one, as used in one of the opening phrases  of the article:

\begin{quote}
North American workers in the logging industry who perform the initial harvesting and transport of trees for ultimate processing into forest products.". 
\end{quote}

\noindent
A fashion-related meaning  is  mentioned and elaborated on later, and it is here that the word \emph{beard} co-occurs with it:

\begin{quote}
a $\cdots$ man who has adopted style traits typical of a traditional lumberjack, namely a beard, plaid shirt,  $\cdots$. 
\end{quote}

Putting these frequencies together, we obtain the following vector representation for it (first row of the following table). 

\begin{center}
\begin{tabular}{|c|c|c|c|c|c|c|c|c|c|c|}
\hline
& pawn & bank & furniture & log & wood & saw & tree & shirt & boot & beard\\
\hline
\hline
&&&&&&&&&&\\
lumberjack & 0 & 0 & 0 & 50 & 8  & 12 & 21 & 2 & 2 & 2\\
&&&&&&&&&&\\
lombard & 16 & 26 & 0&0&0&0&0&0&0&0\\
&&&&&&&&&&\\

\hline
\end{tabular}
\end{center}

\begin{figure}[h]
\begin{center}
\begin{minipage}{10cm}
\begin{center}
\setlength{\unitlength}{0.7mm}
\begin{picture}(60, 70)
  \linethickness{0.3mm}
  \put(26,57){\mbox{\bf wood}}
  \put(30, 20){\vector(1, 0){45}}
  \put(77,18){\mbox{\bf fashion}}
  \put(30, 20){\vector(0, 1){35}}
  \put(30, 20){\vector(-1, -1){25}}
  \put(-4,-12){\mbox{\bf bank}}
   
  \linethickness{0.3mm}
   
  \put(30, 20){\begin{color}{blue}\vector(1, 2){14}\end{color}}
  \put(45,50){\mbox{\bf \begin{color}{blue}lumberjack\end{color}}}
  
%  \put(30, 20){\begin{color}{blue}\vector(1, 2){12}\end{color}}
%  \put(50,40){\text{\bf \begin{color}{blue}dog\end{color}}}
  
  \put(30, 20){\begin{color}{red}\vector(-2, -1){25}\end{color}}
  \put(-15,10){\mbox{\bf \begin{color}{red}lombard\end{color}}}
  
%  \put(30, 20){\begin{color}{red}\vector(-4, -1){25}\end{color}}
%  \put(-15,6){\text{\bf \begin{color}{red}murder\end{color}}}

%  \put(30, 20){\begin{color}{magenta}\vector(-1, 2){14}\end{color}}
%  \put(-17,50){\text{\bf \begin{color}{magenta}gaze $\otimes$ gesture\end{color}}}
 \end{picture}
\end{center}

\vspace{0.9cm}
\end{minipage}
\end{center}
\caption{A  3-Dimensional Semantic Vector Space}
\label{fig:vectspace}
\end{figure}
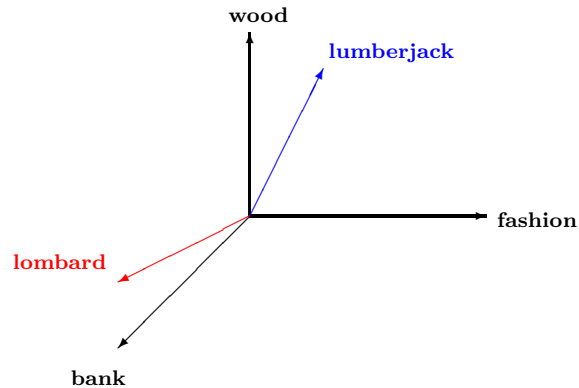

\noindent
We have added to the table co-occurrences of the word \emph{lombard} from  the BNC and in different Wikipedia entry articles for it. Sadly, the word \emph{Langobardi} did neither occur in BNC nor did it have an entry in Wikipedia.  The above co-occurrences are, however, enough to compute a difference in meaning for these two words. If we cluster the columns of the  preceding table in three groups of \emph{wood-related}, \emph{fashion-related}, and \emph{bank-related} meanings, we can plot the vector representations of \emph{Lumberjack} and \emph{Lombard} in the 3-dimensional vector space  shown  in Figure \ref{fig:vectspace}. 

The above vector space shows that \emph{lumberjack} is an ambiguous word, as it has two different meanings: one is its fashion-related meaning and the other is its wood-related meaning. We can use our vector space semantics of pregroup grammars to disambiguate between these meanings.  The procedure is as follows:  if  we modify the word \emph{lumberjack}  with an adjective such as \emph{red} or \emph{flannel}, the resulting vector will get transformed to a vector  closer to the \emph{wood}-related basis. If we modify it with an adjective such as \emph{tall} or \emph{strong}, then the resulting vector will get transformed to a vector closer to the \emph{fashion}-related  basis.

\begin{figure}[h]
\begin{center}
\begin{minipage}{10cm}
\begin{center}
\setlength{\unitlength}{0.7mm}
\begin{picture}(60, 70)
  \linethickness{0.3mm}
  \put(26,57){\mbox{\bf wood}}
  \put(30, 20){\vector(1, 0){45}}
  \put(77,18){\mbox{\bf fashion}}
  \put(30, 20){\vector(0, 1){35}}
  \put(30, 20){\vector(-1, -1){25}}
  \put(-4,-12){\mbox{\bf bank}}
   
  \linethickness{0.3mm}
   
  \put(30, 20){\begin{color}{blue}\vector(1, 2){14}\end{color}}
  \put(45,50){\mbox{\bf \begin{color}{blue}lumberjack\end{color}}}
  
  \put(30, 20){\begin{color}{blue}\vector(2, 1){30}\end{color}}
  \put(63,35){\mbox{\bf \begin{color}{blue}red lumberjack\end{color}}}
  
\thicklines
\put(43,43){\vector(1,-1){10}}

  \put(48,40){\mbox{\bf f}}

  \put(30, 20){\begin{color}{red}\vector(-2, -1){25}\end{color}}
  \put(-15,10){\mbox{\bf \begin{color}{red}lombard\end{color}}}
  
%  \put(30, 20){\begin{color}{red}\vector(-4, -1){25}\end{color}}
%  \put(-15,6){\text{\bf \begin{color}{red}murder\end{color}}}

%  \put(30, 20){\begin{color}{magenta}\vector(-1, 2){14}\end{color}}
%  \put(-17,50){\text{\bf \begin{color}{magenta}gaze $\otimes$ gesture\end{color}}}
 \end{picture}
\end{center}

\vspace{0.9cm}
\end{minipage}
\end{center}
\caption{Transforming  \emph{Lumberjack} to \emph{Red Lumberjack}.}
\label{fig:adjvectspace1}
\end{figure}
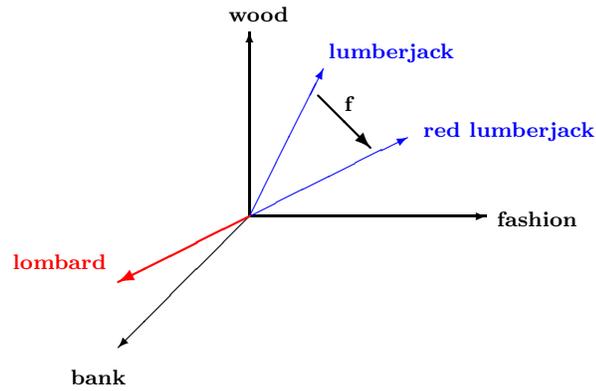

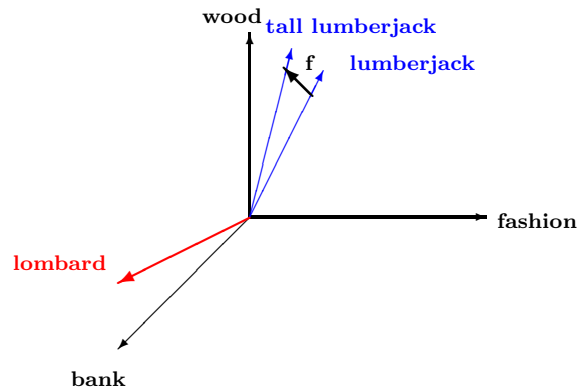
\begin{figure}[h]
\begin{center}
\begin{minipage}{10cm}
\begin{center}
\setlength{\unitlength}{0.7mm}
\begin{picture}(60, 70)
  \linethickness{0.3mm}
  \put(21,57){\mbox{\bf wood}}
  \put(30, 20){\vector(1, 0){45}}
  \put(77,18){\mbox{\bf fashion}}
  \put(30, 20){\vector(0, 1){35}}
  \put(30, 20){\vector(-1, -1){25}}
  \put(-4,-12){\mbox{\bf bank}}
   
  \linethickness{0.3mm}
   
  \put(30, 20){\begin{color}{blue}\vector(1, 2){14}\end{color}}
  \put(49,48){\mbox{\bf \begin{color}{blue}lumberjack\end{color}}}
  
  \put(30, 20){\begin{color}{blue}\vector(1, 4){8}\end{color}}
  \put(33,55){\mbox{\bf \begin{color}{blue}tall lumberjack\end{color}}}
  
\thicklines
\put(42,43){\vector(-1,1){5.5}}

\put(40.5,48){\mbox{\bf f}}

  \put(30, 20){\begin{color}{red}\vector(-2, -1){25}\end{color}}
  \put(-15,10){\mbox{\bf \begin{color}{red}lombard\end{color}}}
  
%  \put(30, 20){\begin{color}{red}\vector(-4, -1){25}\end{color}}
%  \put(-15,6){\text{\bf \begin{color}{red}murder\end{color}}}

%  \put(30, 20){\begin{color}{magenta}\vector(-1, 2){14}\end{color}}
%  \put(-17,50){\text{\bf \begin{color}{magenta}gaze $\otimes$ gesture\end{color}}}
 \end{picture}
\end{center}

\vspace{0.9cm}
\end{minipage}
\end{center}
\caption{Transforming  \emph{Lumberjack} to \emph{Tall Lumberjack}.}
\label{fig:adjvectspace2}
\end{figure}

We work in the rudimentary pregroup grammar of the semantic section, where the set of basic types has two elements $n$ and $s$. We assign the reduced 3-dimensional vector space over the field of reals to the basic type $n$ of our pregroup grammar, and refer to it by $\mathbf{N}$, recall that 
\[
\mathbf{N} \cong  \mathbf{R}^3
\]
 \emph{Lumberjack} has type $n$ and thus its semantic vector is an element of this 3-dimensional space, i.e.  $\ov{\mbox{\it lumberjack}}\in \mathbf{N}$, as depicted in Figure \ref{fig:vectspace}. Adjectives \emph{red} and \emph{tall} have type $n \cdot n^l$ and thus their semantics are tensors in the space $\mathbf{N} \otimes \mathbf{N}^*$, and here we have:
 \[ 
 \mathbf{N} \otimes \mathbf{N}^* \cong \mathbf{R}^9
% \footnote{PHIL:  NO! Tensor products multiply dimensions, so it's $\cong \mathbf{R}^9$}
 \]
 Thanks to the tensor-hom isomorphism $\mathbf{N} \otimes \mathbf{N}^* \cong \mathbf{N} \to \mathbf{N}$,  adjectives can equivalently be seen as linear maps $f_{\mbox{\it red}}, f_{\mbox{\it tall}}$ with type $\mathbf{N} \to \mathbf{N}$ that transform their input vectors in $\mathbf{N}$ to output vectors in $\mathbf{N}$. Figure \ref{fig:adjvectspace1} depicts the transformation corresponding to  $f_{\mbox{\it red}}$.
\[
f_{\mbox{\it red}}(\ov{\mbox{\it lumberjack}}) = \ov{\mbox{\it red lumberjack}}
\]
This transformation decreases the angle between its resulting vector  and the  \emph{fashion}-related basis, demonstrated  by measuring the cosines between each vector and the  basis and verifying the following:
\[
\cos(\ov{\mbox{\it red lumberjack}}, \ov{\mbox{\it fashion}}) \ \geq 
\cos(\ov{\mbox{\it  lumberjack}}, \ov{\mbox{\it fashion}})
\]
The role of adjective $f_{\mbox{\it tall}}$ is the other way around: it transforms $\ov{\mbox{\it  lumberjack}}$ to a vector that is further away from the  \emph{fashion} basis:
\[
f_{\mbox{\it tall}}(\ov{\mbox{\it lumberjack}}) = \ov{\mbox{\it tall lumberjack}}
\]
\[
\cos(\ov{\mbox{\it tall lumberjack}}, \ov{\mbox{\it fashion}}) \ \leq 
\cos(\ov{\mbox{\it  lumberjack}}, \ov{\mbox{\it fashion}})
\]

The compositional vector space semantics, enables us to compute a semantics for all adjective noun phrases and expand our original semantic table, as depicted in 
the chart below: 

\begin{center}
\begin{tabular}{|c|c|c|c|}
\hline
& bank &  wood &fashion\\
\hline
\hline
lumberjack & 0 & 91&  6\\
lombard & 16 & 26 & 0\\
red lumberjack & 0 & 16 & 73\\
tall lumberjack & 0 & 98 & 2\\
\hline
\end{tabular}
\end{center}

The numbers for the new entries are made up, I am assuming that the adjective \emph{red} will have a transformation that numerically transforms the vector of \emph{lumberjack} to somewhere closer to the \emph{fashion}-related  basis, where as the the adjective \emph{tall} will transform the same vector to somewhere close to the \emph{wood}-related  basis, e.g. :
\[
f_{\mbox{\it red}}(0, 91, 6) = (0, 16, 73) \qquad
f_{\mbox{\it tall}}(0, 91, 6) = (0, 98, 2) 
\]
In practice, these  linear maps  are learnt by machine learning the co-occurrence vectors of the nouns they have modified and the co-occurrence  vectors of their holistic adjective noun phrases. For instance, suppose that the holistic vector of the phrase $\ov{\mbox{\it red lumberjack}}$ has occurred with \emph{bank}-related words 0 times, with wood-related words 16 times, and with \emph{fashion}-related words 73 times. Then a linear regression algorithm can use the vector $(0, 91, 6) $ of \emph{lumberjack} and  $(0, 16, 73)$  of \emph{red lumberjack} to learn a linear map that transforms the former to the latter.  If the adjective \emph{red} only ever modified one noun and that noun was \emph{lumberjack}, this transformation just sends 0 to 0, 91 to 16, and 6 to 73. But if we have more than one noun, that is, we have many nouns modified by the adjective red, e.g.   \emph{red car, red carpet, red apple}, then an approximation machine learning algorithm such as linear regression can be used. For details of one such implementation, see \cite{BaroniZam}.  

Sentence types, can in principle get a vector space semantics in any space $\mathbf{S}_{\mathbf{R}}$. It might, however, be useful to keep the analogy with the set theoretic semantics and assign truth values to them. This can be done by sending the basic type $s$ of a sentence to a 2-dimensional space, so we are stipulating the following:
\[
\semantics{s} = \mathbf{R}^2
\]
We interpret the $x$-basis of this space as \emph{true} and its $y$-basis as  \emph{false}. We are not restricted to two discrete truth values and  can, instead of only considering the standard basis as  being true $\left (\begin{array}{c}1\\0\end{array}\right)$  and being false $\left (\begin{array}{c}0\\1\end{array}\right)$, consider two ranges as \emph{degrees of truth} and \emph{degrees of falsity}. In this interpretation, semantics of a sentence becomes a vector with amplitudes on both dimensions $a \left (\begin{array}{c}1\\0\end{array}\right) + b \left (\begin{array}{c}0\\1\end{array}\right)$. This vector encodes the fact that the sentence is true to degree $a$ and false to degree $b$. For instance, for a half true and half false sentence will have $a = b = 1/2$, a fully true sentence will have $a = 1$ and $b=0$ and so on. 

Consider the sentence ``Lumberjacks drink.", in the rudimentary pregroup grammar, the  reduction of this sentence is as follows:
\[
\begin{array}{cccc}
\mbox{Lumberjacks} & \mbox{drink}. &&\\
n & n^r \cdot s & \leq & s
\end{array}
\]
Semantically, \emph{lumberjacks} is a vector in $\mathbf{N}$, \emph{drink} is a tensor in the space $\mathbf{N}^* \otimes \mathbf{S}$. Applying the tenor-hom isomorphism, we think of it as the linear map $f_{\mbox{\it drink}} \colon \mathbf{N} \to \mathbf{S}$. Meaning of ``Lumberjacks drink" is computed by applying this map to the vector of \emph{lumberjack}, as follows:
\[
\ov{\mbox{\it Lumberjacks drink}} = f_{\mbox{\it drink}} (\ov{\mbox{\it  lumberjack}}) \quad \in \mathbf{S}
\]
This  results in a vectors in $\mathbf{S}$. In order to exemplify what this vector might denote, let us consider our two adjective noun phrases. The sentence with the phrase \emph{tall lumberjacks} as its subject  is quite plausible, whereas the sentence with the phrase \emph{red lumberjack} as its subject  is less so. We assign the truth value 1 to the plausible cases and 0 to the non-plausible ones and compute:
\begin{eqnarray*}
\ov{\mbox{\it Tall lumberjacks drink}} &=& f_{\mbox{\it drink}} ( f_{\mbox{\it tall}}(\ov{\mbox{\it  lumberjack}})) = \left ( \begin{array}{c} 1\\0\end{array}\right)\\
\ov{\mbox{\it Red lumberjacks drink}} &=& f_{\mbox{\it drink}} ( f_{\mbox{\it red}}(\ov{\mbox{\it  lumberjack}})) = \left ( \begin{array}{c} 0\\1\end{array}\right)
\end{eqnarray*}

It is possible to work with degrees of truth and soften the above extreme interpretations to the following ones:

\begin{eqnarray*}
\ov{\mbox{\it Tall lumberjacks drink}} &=& f_{\mbox{\it drink}} ( f_{\mbox{\it tall}}(\ov{\mbox{\it  lumberjack}})) =  0.8 \left ( \begin{array}{c} 1\\0\end{array}\right) +  0.2 \left ( \begin{array}{c} 0\\1\end{array}\right)\\
\ov{\mbox{\it Red lumberjacks drink}} &=& f_{\mbox{\it drink}} ( f_{\mbox{\it red}}(\ov{\mbox{\it  lumberjack}})) =  0.3\left ( \begin{array}{c} 1\\0\end{array}\right) +  0.7 \left ( \begin{array}{c} 0\\1\end{array}\right)\\
\end{eqnarray*}

In either of these interpretations, and similar to what we discussed in the adjective noun case, one can learn the linear map $f_{\mbox{\it drink}}$ corresponding to the verb \emph{drink}.  Similar machine learning algorithms can be employed here, e.g. for a multi-linear regression algorithm used to learn verbs see \cite{MultiStep}, for an improvement on it, see \cite{Tamara}. 

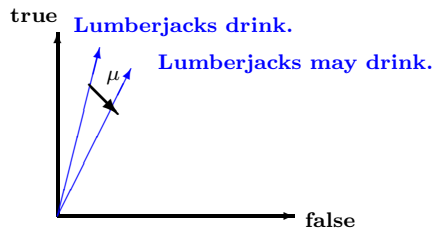
\begin{figure}[h]
\begin{center}
\hspace{-3cm}
\begin{minipage}{10cm}
\begin{center}
\setlength{\unitlength}{0.7mm}
\begin{picture}(60, 70)
  \linethickness{0.3mm}
  \put(21,57){\mbox{\bf true}}
  \put(30, 20){\vector(1, 0){45}}
  \put(77,18){\mbox{\bf false}}
  \put(30, 20){\vector(0, 1){35}}

  \linethickness{0.3mm}
   
  \put(30, 20){\begin{color}{blue}\vector(1, 2){14}\end{color}}
  \put(49,48){\mbox{\bf \begin{color}{blue}Lumberjacks may drink.\end{color}}}
  
  \put(30, 20){\begin{color}{blue}\vector(1, 4){8}\end{color}}
  \put(33,55){\mbox{\bf \begin{color}{blue}Lumberjacks  drink.\end{color}}}
  
\thicklines
\put(36,45){\vector(1,-1){5.5}}

\put(39.3,46){$\mu$}

%  \put(30, 20){\begin{color}{red}\vector(-4, -1){25}\end{color}}
%  \put(-15,6){\text{\bf \begin{color}{red}murder\end{color}}}

%  \put(30, 20){\begin{color}{magenta}\vector(-1, 2){14}\end{color}}
%  \put(-17,50){\text{\bf \begin{color}{magenta}gaze $\otimes$ gesture\end{color}}}
 \end{picture}
\end{center}

\vspace{0.9cm}
\end{minipage}
\end{center}
\caption{Transforming the meaning of a sentence.}
\label{fig:may}
\end{figure}

For a slightly more complex example, consider the sentence ``Lumberjacks may drink". In order to work out a reduction for this sentence, we need to elaborate our pregroup grammar with new basic types $\pi_3, s_1, i, j$. Recall, however, that we had the partial orderings $\pi_i \leq n$ and $i \leq j, s_i \leq s$. So we  assign the same vector space $\mathbf{N}$ to the type $\pi_3$ and the same vector space $\mathbf{S}$ to the type $s_1$. For $i$ and $j$, we stipulate  $\semantics{i} = \semantics{j}$. On the syntactic side, all these types say is that  they stand for infinitives of verbs and their semantics is the same as the semantics of that verb. Hence, on the semantics side, we need to use the type of  the verb for them. If we do not, the auxiliary \emph{may} will be tasked to assign a meaning to the verb \emph{drink}, but it  occurs before any  verb what so ever, e.g. as in  `` Lumberjacks may sleep.", or ``Lumberjacks may sneeze" and thus it cannot represent the meanings of all the verbs that it  modifies. So we assign the semantic type $N^* \otimes S$ to $i$ and $j$. As a result,  semantics of \emph{may} becomes an element of the following tensor space: 
\[
N^* \otimes S \otimes (N^* \otimes S)^*
\]
By the tensor-hom isomorphism, we can equivalently think of it as the following linear map
\[
f_{\mbox{\it may}} \colon  (N^* \otimes S) \to (N^* \otimes S)
\]
That is, a map that transforms a verb into another verb. The meaning of our sentence thus becomes as follows:
\[
f_{\mbox{\it may}} (\ov{\mbox{\it Lumberjacks}}, f_{\mbox{\it drink}}) 
\]
The semantic role of \emph{may} is to act as an evaluation map and apply its verb input to its noun input. But it also changes the truth value of the resulting sentence, from  more  true to less so. So we suppose it is a composition of an $ev$ map with an endomorphism $\mu \colon S \to S$, that is
\[
f_{\mbox{\it may}} := \mu \circ ev
\]
Instantiating this in the above will provide us with the following as the meaning  of our sentence:
\[
=  \mu \circ ev(\ov{\mbox{\it Lumberjacks}}, f_{\mbox{\it drink}}) = \mu(f_{\mbox{\it drink}}(\ov{\mbox{\it Lumberjacks}}))
\]
Suppose that  the meaning of ``Lumberjacks drink" is  $0.8 \left ( \begin{array}{c} 1\\0\end{array}\right) +  0.2 \left ( \begin{array}{c} 0\\1\end{array}\right)$, then the $\mu$ map will nudge it more towards the false axis:
\[
\mu(0.8 \left ( \begin{array}{c} 1\\0\end{array}\right) +  0.2 \left ( \begin{array}{c} 0\\1\end{array}\right)) = 0.75 \left ( \begin{array}{c} 1\\0\end{array}\right) +  0.35 \left ( \begin{array}{c} 0\\1\end{array}\right)
\]
So the meaning of ``Lumberjacks may drink" becomes less true than the meaning of ``Lumberjacks drank", as shown in Figure \ref{fig:may}. 

We bring this section to an end by analysing questions. In computational linguistics, the meaning of a question is sometimes taken to be the sentence about which it is asked and which  provides a suitable answer to it.  For instance, we can ask three questions about the sentence ``Lumberjacks may drink":  a yes-no question ``May lumberjacks drink?" and two wh-questions:  ``Who may drink"  and ``What may lumberjacks do?".  The answer to all of these questions is the sentence about which they are asked: ``Lumberjacks may drink".  Accordingly, we assign the semantic space $\mathbf{S}$ to the types $q_1$ and $\bar{q}$ in order to compute the semantics of these questions. Recall that in a yes-no question, \emph{may} had type $q_1 i^l \pi^l$; its semantic type becomes as follows, taking $\semantics{i}$ and $\semantics{\pi}$ to be the same as before:
\[
S \otimes (N^* \otimes S)^* \otimes N^*
\]
It is easy to verify that the semantics of ``May lumberjacks drink?" becomes the same as the semantics of ``Lumberjacks may drink", that is:
\[
\mu(f_{\mbox{\it drink}}(\ov{\mbox{\it Lumberjacks}}))
\]
In the who-question, the syntactic type of \emph{who} was $\bar{q}s_1^l \pi_3$; its semantic type becomes as follows:
\[
S \otimes S^* \otimes N
\]
which is equivalent to
\[
S \otimes (N^* \otimes S)^* \cong Hom((N^* \otimes S), S)
\]
The auxiliary \emph{may}  retains the same type as it had in the declarative sentence, its semantic type will also remain as before, that is $N^* \otimes S \otimes (N^* \otimes S)^*$. Concretely, it is a linear map of the following type
\[
f_{\mbox{who}} \colon (N^* \otimes S) \to S
\]
where it takes a verb  in $N^* \otimes S$ and returns a sentence about that verb, i.e. a sentence which has that verb as its verb. This sentence is the answer to the wh-question. 

In order to compute the semantics of the question, we first apply $f_{\mbox{may}}$ to $f_{\mbox{drink}}$, then apply $f_{\mbox{who}}$ to the result and obtain a vector for it in the sentence space:
\[
f_{\mbox{who}}(f_{\mbox{may}} (f_{\mbox{drink}})) \in \mathbf{S}
\]
 In our example,  the verb \emph{drink}   is passed to  $f_{\mbox{who}}$  upon receipt  of which, it outputs a sentence with \emph{drink} as its verb, i.e.  the sentence ``Lumberjacks may drink", which also serves as its answer. Note that any other sentence with \emph{drink} as its verb would also be acceptable as an answer to this question and our method takes this into account. 
 
 \section{What is  Truth?}
 \label{truth}
 
We  showed how to develop a vector space semantics for pregroup grammars, and   that this semantics, as opposed to the set theoretic one, is unambiguous. One can however, worry about the notion of \emph{truth}  this semantics offers. Vectors are arrays of numbers, truth and falsity are constants. Has our quest for an unambiguous semantics costed us the  safety of having a notion of  \emph{truth}? Sadly, at the moment of writing this paper, the answer to this question seemed unclear.  

Passages between vector space and set theoretic semantics,  through a  finite model of first order logic and tensors of vector spaces have been explored in \cite{grefenstette2013,rocktaschel2015,Sato17,HedgesSadr2019} and might become of use.  A pragmatic approach,  considered in previous work \cite{Coeckeetal,APAL,Maillard2014},  fixed the semantics of sentences, i.e. $\semantics{s}$, to be the two dimensional space $\mathbf{k}^2$, which was referred to as a \emph{plausibility space}  and explored in  \cite{Clark2013}.  A method for constructing plausibility spaces from corpora of data was  implemented and experimented with  in \cite{Tamara}. In \cite{Sadrzadehetal}., we demonstrated. via examples, how a two dimensional plausibility space can be reduced to a one dimensional one, namely the real line i.e. the vector field $\mathbf{k}$.  

As our reviewer suggested, however, generalising the set theoretic notion of  truth, according to the Curry-Howard-Lambek isomorphism, to this two dimensional space, or any  finite dimensional space $\mathbf{k}^n$ is not trivial. An option is to work with the  Linear  L\"{a}uchli  semantics,  developed for Linear Logic  \cite{BluteScott96}, where one interprets the grammatical reduction, by induction. In our case, a grammatical reduction, which is a pregroup partial ordering  $\semantics{t_1,\ldots,t_n\leq s}$, is   interpreted as the homset (i.e. linear maps)
$Hom(\semantics{t_1}\otimes \cdots \otimes \semantics{t_n}, \semantics{s})$.  This is defined to be true when it contains a canonical element, e.g. the constant map 1 on the set $\semantics{s} = \{0,1\}$ for sets, projections into the standard basis for vector spaces for .$\semantics{s} = \mathbf{k}^n$ In the latter case, however,  one will be confronted  with many, rather than just one,  notions of truth, and one canonical falsity, namely the origin. Working out the details of this approach is a natural future direction of this paper.

\section{Conclusion}
\label{sec:conclusion}
We  reviewed the syntax and ambiguous set theoretic semantics of pregroup grammars. We then followed a suggestion of Lambek and developed  a vector space semantics for pregroups. We showed how using direct sum of vector spaces, as seemingly indicated in the notation used by Lambek,  does still lead to ambiguity, but if dimensionality explosion is tolerated and direct sum is replaced by tensors,  the problem gets resolved. On the practical side, we build semantic vector representations for some exemplary words, phrases, and sentences of language and show how compositionality of  vector semantics  disambiguates meaning. Finally, for the first time, we present a vector semantics for questions and show how their representations become the same as the sentences they are asked about. 

Overall, until very recently, only the basic fragment of English, consisting of Subject-Verb-Object sentences with adjectival modifiers were considered in a \emph{DisCoCat}. Recently, relative pronouns \cite{RelPronMoL,SadrClarkCoecke1,SadrClarkCoecke2} and quantifiers \cite{HedgesSadr2019}  were also formalised. A preliminary study of vector space semantics for sentences with the discourse phenomena, e.g.  VP-ellipsis with anaphora,  has also been pursued in the work done by  my PhD student Gijs Wijnholds \cite{WijnholdsSadr2019,WijnholdsSadrNAACL} towards his dissertation. There is some other relevant work in this area, but not directly via pregroups or vectors, e.g.  for knowledge bases  \cite{CoeckeToumi} and using Dynamic Syntax \cite{Sadrzadehetal}.

In order to analyse these complex phenomena of natural language and the phenomena arising in mildly context sensitive languages such as Dutch and Swiss-German, in Chapter 7 of \cite{Lambek2008}, Lambek argues that pregroups may be amended with lattice operations and/or products of them with themselves should be considered. Exploring these phenomena and developing appropriate vector space semantics for them is work in progress.

\bibliography{ref.bib}
\bibliographystyle{plain}

\end{document}